# Novel Language Resources for Hindi: An Aesthetics Text Corpus and a Comprehensive Stop Lemma List

Gayatri Venugopal-Wairagade[1], Jatinderkumar R. Saini[2], Dhanya Pramod[3]
Symbiosis Institute of Computer Studies and Research, Symbiosis International (Deemed University), Pune, India[1, 2]
Symbiosis Centre for Information Technology, Symbiosis International (Deemed University), Pune, India[3]

***Abstract*—This paper is an effort to complement the contributions made by researchers working toward the inclusion of non-English languages in natural language processing studies. Two novel Hindi language resources have been created and released for public consumption. The first resource is a corpus consisting of nearly thousand pre-processed fictional and non-fictional texts spanning over hundred years. The second resource is an exhaustive list of stop lemmas created from 12 corpora across multiple domains, consisting of over 13 million words, from which more than 200,000 lemmas were generated, and 11 publicly available stop word lists comprising over 1000 words, from which nearly 400 unique lemmas were generated. This research lays emphasis on the use of stop lemmas instead of stop words owing to the presence of various, but not all morphological forms of a word in stop word lists, as opposed to the presence of only the root form of the word, from which variations could be derived if required. It was also observed that stop lemmas were more consistent across multiple sources as compared to stop words. In order to generate a stop lemma list, the parts of speech of the lemmas were investigated but rejected as it was found that there was no significant correlation between the rank of a word in the frequency list and its part of speech. The stop lemma list was assessed using a comparative method. A formal evaluation method is suggested as future work arising from this study.**

## I. Introduction

One of the basic requirements to devise a tool to perform any task in Natural Language Processing (NLP) is a corpus that represents the target language or the target domain. Adhering to the ‘Bender Rule’, according to which researchers are required to name the language that was targeted by the study, we would like to inform the readers that the study focuses on Hindi, the official language of India. The language ranks third on the list of the languages with the largest number of first language speakers in the world [1]. The study aims at building a corpus in the aesthetics domain and utilizing the corpus, along with other corpora to publish an exhaustive list of stop lemmas based on their raw frequencies in the corpora. The corpus has been released under the GNU General Public License for public use, in order to ease the process of corpus acquisition. It is also necessary to mention that the text and the corpora that were acquired from various sources have been used solely for academic and research purpose. The corpus was created because of the difficulty that we faced while searching for and acquiring novels, stories and non-fictional content written by contemporary authors as well as content written by authors in India’s pre-independence era, i.e., prior to 1947. The broad objective of the study is to utilize and release unbiased time-independent text in the form of a corpus. Since the study emphasizes the lexical aspect of text, features related to context and discourse have not been discussed in this paper. The stop lemma list created as part of this work has been built using text from multiple sources, thus making it suitable for generic consumption. The reported outcome is best as on date subject to the data used for this research. We believe that this corpus and the list of stop lemmas would be a useful resource for NLP tasks such as creating language models, text classification and information retrieval in Hindi.

The remaining paper is organized into three sections. Section II of the paper consists of the existing work in this area and the research questions. Section III contains the description of the corpus and the methodology along with the final list of stop lemmas. Section IV consists of the concluding remarks.

## II. Existing Work and Objectives

Stop words are words that are present in a sentence solely for grammatical reasons and do not contribute to the information obtained from the text [2]. Hence if these stop words are identified and removed before using the text for a task, the performance of the task could improve. Many studies have focused on the importance of removing stop words as a pre-processing step for text processing tasks [3], [4] and [5]. Author in [6] manually extracted stop words based on parts of speech such as pronouns, prepositions, conjunctions etc. from two news-based corpora to create a stop word list in Hindi consisting of 275 words. Author in [7] created a stop word list by converting words to lemmas in a corpus of news articles consisting of 441,153 words. Author in [8] proposed a method for automatic stop word generation that created stop word lists that matched the top twenty stop word lists from four publicly available lists. Their corpus was based on news articles. Although we found studies that focused on automatic stop word generation [9], [10], [11], [12], [13] and others that published the stop word lists for public use [14], we could not find an exhaustive publicly available list of stop words in Hindi based on multiple corpora. Another problem with multiple lists based on one or two corpora is the inconsistency of the words in the lists [15]. Author in [16] manually created a stop word list consisting of 256 words from Punjabi poetry and articles. This list was brought down to 184 unique words by lemmatizing the words. We would like to emphasize here that the researchers lemmatized the words after identifying a word as a stop word as opposed to lemmatizing all the words and then identifying the stop words. [17] collated a list of stop words in numerous languages spanning multiple countries.

The researchers observed the presence of multiple word forms in the stop word lists depending on the morphological complexity of the language. We catered to this issue while framing the objectives of our study. Author in [18] manually created a list of stop words based on news corpora in Gujarati. The researchers also aimed to find a pattern in the stop words that were assigned the same part of speech tag but they could not find any similarity. Another study used a threshold based approach to identify stop words in Sanskrit text by calculating the frequencies of the words in the text [19]. The researchers removed nouns from the list under the assumption that nouns cannot be treated as stop words. We aimed to test the validity of this assumption through our study.

The source of data in majority of the studies discussed in this section was news articles. All these studies, except for one, focused on the creation of stop word lists, instead of stop lemma lists. This would lead to the creation of a list that consists of a morphological variation of a word instead of the root word itself, which could negatively affect the quality of the NLP task owing to the non-robust nature of the list, hence defeating the purpose of its creation. One of the studies focused on assigning a part of speech tag depending on the stop word. We formed one of our objectives along similar lines, but our emphasis was on the relationship between parts of speech and stop words in general, rather than on categorizing similar stop words under a particular lexical category.

The objectives of the study were framed as follows, based on the analysis of the existing work:

*a)* To determine whether the existing publicly available stop word lists are inter-changeable

*b)* To determine whether the part of speech of a word can be used to determine whether the word is a stop word or not

*c)* To create an exhaustive stop lemma list irrespective of the target domain

In order to meet the above objectives, our study was focused on finding answers to the following research questions:

RQ1: Are the top ten stop words across all the available stop words lists the same, irrespective of the rank of the words in the lists?

RQ2: Is there any significant relationship between:

*a)* The part of speech of a word and its rank in the stop word list?

*b)* The part of speech of a lemma and its rank in the stop lemma list?

RQ3: How can the available resources be synthesized to create an exhaustive stop lemma list?

## III. Methodology

### A. Corpus Creation and Corpora Metadata

This section discusses the creation of our corpus and throws light on the metadata of the corpora we collected. We scraped novels and stories from http://hindisamay.com, an e-library maintained by Mahatma Gandhi Antarrashtriya Hindi Vishwa Vidyalaya (translated to Mahatma Gandhi International Hindi University), Wardha, http://premchand.co.in, a website dedicated to the popular novelist Premchand's stories, and Bhandarkar Oriental Research Institute's Digital Library (http://borilib.com). Scrapy, an open source tool, was used to extract content from websites. We also extracted content from PDF of novels that were not available on the specified websites, but the text thus extracted could not be used because of encoding errors. Owing to issues such as this and lack of availability of public content, the size of the corpus is not comparable with that of articles present in Wiki dumps. However, the corpus, in combination with other corpora can become an effective resource.

As a preprocessing step, we split the text into sentences, tokenized the sentences and deleted special characters, English tokens and Latin numbers. Joined words were not segmented. The details with respect to the vocabulary of our corpus and the corpora that we acquired from various sources to create a stop word list, can be seen in Table I. LR indicates the language resource specified in the Language Resources section. LR#12 refers to the corpus that we created.

The numbers presented in the table against LR#13 to LR#14 were generated by calculating the raw frequency of the words, and by fetching the lemmas of the words present in the corpora and calculating the raw frequency of the lemmas, as this information was not readily available. The corpora included in the table span over ten domains and a period of over hundred years.

The state-wise distribution of the authors whose works have been included in the corpus, can be seen in Fig. 1.

As depicted in the figure, majority of the work is associated with authors whose native state is Uttar Pradesh, a state in northern India. Another observation made in the metadata was that only 4.84% of the publications are authored by females. This number was 0 in the pre-independence era for this corpus. However, the number increased as the years passed, although it is incomparable with that of the male authors.

TABLE. I. METADATA OF THE CORPORA THAT ARE PART OF THE STUDY

| S.No. | Source | Unique Word Count | Unique Lemma Count | Domain |
|---|---|---|---|---|
| 1 | LR#12 | 145508 | 118266 | Aesthetics |
| 2 | LR#13 | 21335 | 17159 | Entertainment |
| 3 | LR#14 | 119313 | 102201 | Not available |
| 4 | LR#15 | 2330 | 1851 | Varied domains |
| 5 | LR#16 | 21826 | 18220 | Tourism |
| 6 | LR#17 | 39351 | 32074 | Agriculture and Entertainment |
| 7 | LR#18 | 35018 | 28645 | Agriculture, Entertainment, Politics and Public Administration, Sports, Religion, Literature, Aesthetics, Economy |
| 8 | LR#19 | 20430 | 16673 | Health |

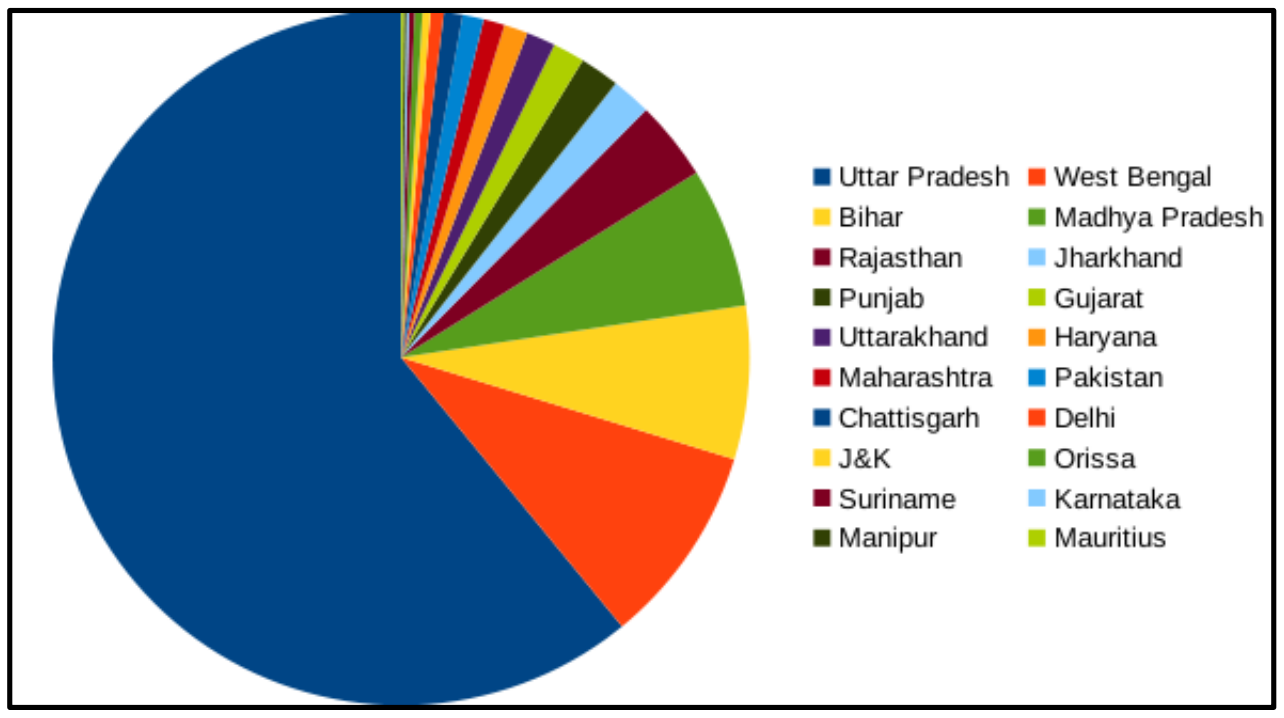

Fig. 1. State-Wise Distribution of Authors in the Corpus.

978 articles including novels, short stories and non-fictional texts were collated from the sources mentioned earlier. Out of these 978 articles, the metadata of 164 articles could not be found. We were unable to find details of authors who are not highly accomplished, but we believe that the authors are not amateur writers since the university chose to publish their work on their website.

## B. *Stop Word List Creation*

In this section, we attempt to answer the research questions framed in Section 2.

RQ1: Are the top ten stop words across all the available stop words lists the same, irrespective of the rank of the words in the lists?

The null hypothesis was framed as follows:

$H_0$: The top ten stop words across all the available stop words lists are the same, irrespective of the rank of the words in the lists.

$H_A$: The top ten stop words across all the available stop word lists are not the same, irrespective of the rank of the words in the lists.

We created a list of the top ten stop words obtained from publicly available lists and generated the list from corpora released by Technology Development for Indian Languages (TDIL), Ministry of Electronics & Information Technology (MeitY), Government of India., Centre for Indian Language Technology (CFILT), IIT Bombay, Wikipedia dump, http://opus.nlpl.eu that consists of subtitles obtained from opensubtitles.org, and the aesthetics corpus that we created using articles collected from the sources mentioned above. We did not have access to the corpora that were used to generate the publically available lists as these lists have neither been released as formal resources nor has the process of creation of these lists been published as research papers. Table II contains the list of the top ten stop words in each source.

A word cloud generated from the list of these stop words and their count in the lists across all the sources, can be seen in Fig. 2. The size of the word in the figure is directly proportional to its count. It was observed that the word lists were not consistent across the sources. Out of the 19 sources and 82 unique stop words, the maximum count of sources in which a word appeared was 14.

TABLE. II. Top Ten Stop Words in Each Source

| Source | Stop Words | | | | | | | | |
|---|---|---|---|---|---|---|---|---|---|
| LR#1 | में | है | हैं | नहीं | लिए | गया | तथा | अपने | कुछ | साथ |
| LR#2 | जैसा | मैं | उसके | कि | वह | था | के लिए | पर | हैं | साथ |
| LR#3 | की | और | एक | तक | में | है | आप | कि | यह | वह |
| LR#4 | के | है | में | की | से | और | का | को | हैं | पर |
| LR#5 | एक | आप | और | यह | कर | हम | वह | पर | इस | अब |
| LR#6 | अत | अपना | अपनी | अपने | अभी | अंदर | आदि | आप | इत्यादि | इन |
| LR#7 | अंदर | अत | अदि | अप | अपना | अपनि | अपनी | अपने | अभि | अभी |
| LR#8 | पर | इन | वह | यिह | वुह | जिन्हें | जिन्हों | तिन्हें | तिन्हों | किन्हों |
| LR#9 | मैं | मुझको | मेरा | तुम्हारा | हमने | हमारा | अपना | हम | आप | आपका |
| LR#10 | के | का | एक | में | की | है | यह | और | से | हैं |
| LR#11 | और | पर | एक | रत | कर | इस | यह | अन | वर | सम |
| LR#12 | है | के | में | की | और | से | का | को | नहीं | तो |
| LR#13 | है | के | मैं | नहीं | में | हैं | एक | आप | और | लिए |
| LR#14 | के | है | में | की | से | और | का | को | हैं | पर |
| LR#15 | के | में | की | और | लिए | हैं | है | से | का | को |
| LR#16 | के | है | में | और | की | से | एक | का | हैं | को |
| LR#17 | के | है | में | की | का | से | और | को | हैं | भी |
| LR#18 | के | है | में | की | का | से | को | और | हैं | ने |
| LR#19 | है | के | में | से | की | को | का | हैं | और | हो |

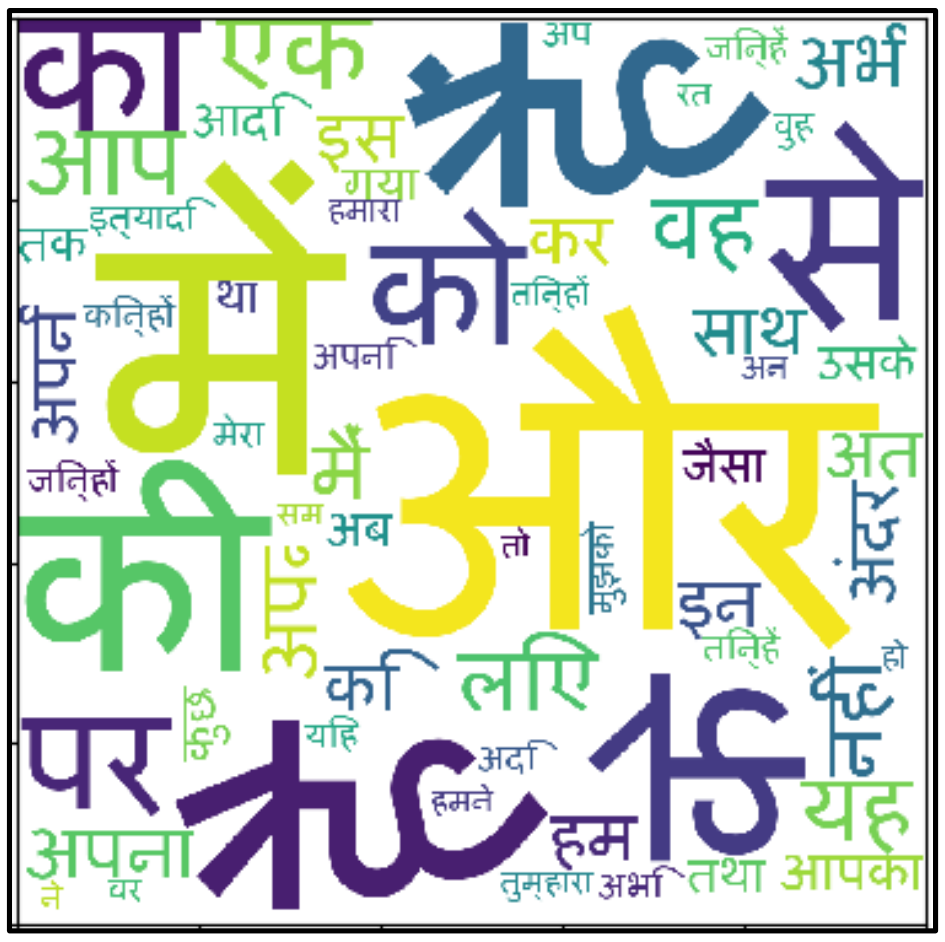

Fig. 2. Word Cloud based on the Count of the Top Ten Stop Words Across 19 Sources.

We also found inconsistencies in the list of lemmas. The list of top ten stop lemmas from each source can be seen in Table III. Table III consists of fewer language resources as compared to Table II, as there exists to publicly available lists of stop lemmas at the time of writing this paper. Hence, we used the corpora in hand to generate the lemmas of the words and sorted them based on their raw frequencies.

A word cloud generated from the list of the top 10 lemmas and their count in this list across 8 sources (LR#12 to LR#19), can be seen in Fig. 3.

An interesting observation here was that among 8 sources and 22 unique stop lemmas, the maximum count of sources in which a lemma appeared was 8, which is a significant improvement over the count of the words in the stop word list. Hence, though we reject our null hypothesis and accept the alternative hypothesis, we believe that while creating an exhaustive list of stop words, we should consider the lemmas of the words and not just various morphological variations of a word.

RQ2: Is there any significant relationship between:

*a)* The part of speech of a word and its rank in the stop word list?

*b)* The part of speech of a lemma and its rank in the stop lemma list?

TABLE. III. TOP TEN STOP LEMMAS IN EACH SOURCE

| Source | Stop Lemmas | | | | | | | | | |
|---|---|---|---|---|---|---|---|---|---|---|
| LR#12 | का | है | वह | हो | में | कर | था | जा | यह | और |
| LR#13 | है | का | मैं | कर | वह | यह | नहीं | हो | में | आप |
| LR#14 | का | है | में | वह | हो | कर | यह | जा | से | और |
| LR#15 | का | है | में | और | कर | जा | से | को | दे | यह |
| LR#16 | का | है | कर | में | और | से | एक | जा | यह | ले |
| LR#17 | का | है | कर | में | यह | हो | जा | से | और | वह |
| LR#18 | का | है | कर | में | यह | हो | वह | से | को | जा |
| LR#19 | है | का | कर | में | हो | से | यह | जा | को | और |

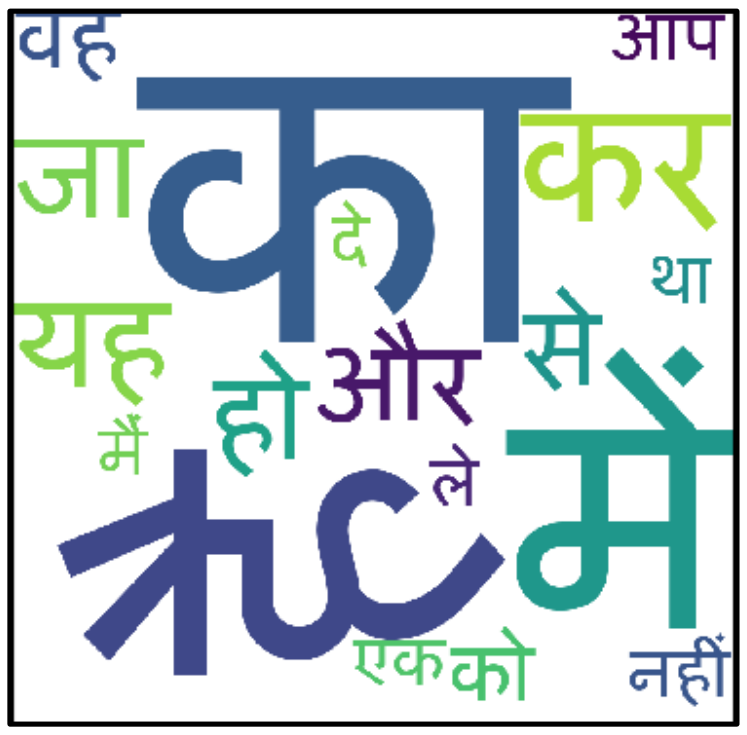

Fig. 3. Word Cloud based on the Count of the Top Ten Stop Lemmas Across 8 Sources.

The null hypotheses were framed based on a general assumption that parts of speech such as prepositions, postpositions and symbols are commonly marked as stop words.

*a)* $H_0$: There is a significant relationship between the part of speech of a word and its rank in the stop word list.

$H_A$: There is no significant relationship between the part of speech of a word and its rank in the stop word list.

*b)* $H_0$: There is a significant relationship between the part of speech of a word and its rank in the stop word list.

$H_A$: There is no significant relationship between the part of speech of a lemma and its rank in the stop lemma list.

In order to test the hypotheses, we considered the parts of speech of all the words that appeared in the top ten ranks of the stop word lists for each source from S12 to S19. The parts of speech for every word and every lemma in each list was determined for all the sources using the part of speech generated from the Hindi treebank of the stanfordnlp POS processor. The point biserial correlation coefficient between the parts of speech present in the top ten ranks of the stop words lists, and the rank of the word in the list was calculated. The parts of speech of 10722288 unique words and 335089 lemmas were analyzed. The results can be seen in Table IV. The table consists of descriptive statistics of the coefficient values across the lists generated from all the sources.

It was observed that there was no significant relationship between any part of speech and the words' ranks. Hence the null hypotheses were rejected and the alternative hypotheses were accepted. This bursts the assumption that words belonging to parts of speech such as prepositions, postpositions, conjunctions etc. would be ranked among the top stop words.

RQ3: How can the available resources be synthesized to create an exhaustive stop lemma list?

TABLE. IV. DESCRIPTIVE STATISTICS OF THE CORRELATION BETWEEN THE PART OF SPEECH OF THE WORDS AND THEIR CORRESPONDING RANK IN THE LIST OF STOP WORDS

| Part of Speech | Source List | Point Biserial Coefficient Value | | | | P-Value | |
|---|---|---|---|---|---|---|---|
| | | *Mean* | *SD* | *Max* | *Min* | *Mean* | *SD* |
| *NN/NNP/ NNPC* | Words | -0.02 | 0.09 | 0.09 | -0.18 | 0.05 | 0.10 |
| | Lemmas | -0.02 | 0.11 | 0.22 | -0.16 | 0.04 | 0.11 |
| *PSP/PRP* | Words | -0.07 | 0.04 | -0.01 | -0.11 | 0.06 | 0.16 |
| | Lemmas | -0.06 | 0.04 | -0.01 | -0.11 | 0.06 | 0.16 |
| *SYM* | Words | 0.10 | 0.06 | 0.16 | 0.00 | 0.10 | 0.31 |
| | Lemmas | 0.11 | 0.06 | 0.20 | 0.04 | 0.07 | 0.18 |
| *VM* | Words | -0.04 | 0.06 | 0.09 | -0.08 | 0.13 | 0.27 |
| | Lemmas | -0.03 | 0.06 | 0.09 | -0.08 | 0.20 | 0.29 |
| *QC/QF/QO* | Words | -0.05 | 0.02 | -0.02 | -0.08 | 0.05 | 0.12 |
| | Lemmas | -0.06 | 0.03 | -0.02 | -0.10 | 0.04 | 0.13 |
| *NEG* | Words | -0.02 | 0.02 | 0.00 | -0.05 | 0.08 | 0.05 |
| | Lemmas | -0.02 | 0.02 | 0.00 | -0.06 | 0.15 | 0.22 |
| *CC* | Words | -0.04 | 0.02 | -0.01 | -0.07 | 0.08 | 0.24 |
| | Lemmas | -0.04 | 0.02 | -0.01 | -0.07 | 0.08 | 0.22 |

In an attempt to solve this question, we collected the top 100 stop words from the existing publicly available lists, took a union of all the lists, and replaced the words with their lemmas. The lemmas were used owing to the observation made while searching for the answer to RQ1. This was named Set A. Set B was created by taking a union of the top 100 stop lemmas from all the corpora. The exhaustive stop word list C was another set that was generated by taking an intersection of Set A and Set B.

Set A = genLemma($Set_1$) U genLemma($Set_2$) U genLemma($Set_n$),

where $Set_i$ represents the set of stop words in source i; 1<=i<=11 and genLemma is a function that replaces each word in the set with its corresponding lemma

Set B = $Set_1$ U $Set_2$ U ... $Set_m$,

where $Set_j$ represents the set of the top 100 most frequent lemmas in source j; 12<=j<=19

Set C = Set A ∩ Set B

1096 words from publicly available lists were brought down to 1071 words after removing the duplicate words and phrases. A list of unique lemmas of these words was created. The length of this list was 370.

Set B, which was created by fetching the lemmas of all the words in the corpora, consisted of 213554 lemmas. These lemmas were generated from 13811781 words that were brought down to 405111 words after removing the duplicates.

The final list of stop lemmas was created by combining both the lists and extracting the common lemmas. The list consists of 311 lemmas, as can be seen in Table V. The lemmas have been arranged in decreasing order of frequency from top to bottom.

The complete list is available on https://github.com /gayatrivenugopal/hindi-corpus-stoplemmas.

In order to assess this list, we compared our list with the English stop word list present in Python's Natural Language ToolKit package (nltk). We used Google Translate with manual intervention to translate the English stop words into their equivalent Hindi words. We chose English as we could not ascertain the correctness of translation of other languages in Hindi. The lemmas of these translated words in Hindi were generated and compared with the exhaustive stop lemma list. 74 unique equivalent Hindi stop lemmas were generated for the 179 words in the English stop word list. We could not translate the words 'being', 'will' and 'shall'. Although 'shall' and 'will' were not present as it is in the English stop word list, we expanded the word form ''ll' to 'shall' and 'will'. An ambiguity of a particular kind was found while analyzing the words in the English stop words list. The list contained the word 'won', which was next to words such as 'shan' and 'aren'. Here, we did not consider the meaning of 'won' to be the past tense of 'win'. On the other hand, it was a form of wouldn't, i.e., won't, wherein the apostrophe and the 't' were removed from the word. We used the same approach for words such as 'shant', 'aren', etc. Following the disambiguation, translation and lemmatization phase, we observed that out of the 74 unique lemmas, 73 lemmas were present in our stop lemma list. The lemma that was absent from our list was 'जरूर', the Hindi equivalent of the English stop word 'must'.

TABLE. V. 311 STOP LEMMAS PRESENT IN THE EXHAUSTIVE LIST

| | | | | | | |
|---|---|---|---|---|---|---|
| का | लेकिन | बाहर | अथवा | दर | तर | टर |
| है | तरह | पूछ | मत | गलत | कौनसा | यक |
| वह | अब | भारत | पुरुष | खिलाफ | कोन | मह |
| में | बहुत | छोटा | रास्ता | जन | लत | पनी |
| कर | दिन | सामने | जैसे | पालन | निहायत | उनकि |
| हो | रख | बीच | आवश्यकता | आठ | कवर | तथ |
| यह | जब | तीन | कल | जोड़ | बंदरगाह | उनक |
| जा | लगा | हर | भीतर | पल | दुसरा | उत |
| और | तथा | जहाँ | कोशिश | बिलकुल | तिन्ह | वग़ैरह |
| से | बाद | केवल | प्रत्येक | उच्च | बाला | रक |
| था | चाह | डाल | औरत | सदा | तिस | वुह |
| को | यहाँ | कितना | दस | अधिकांश | उह | |
| मैं | दूसरा | बना | खाना | वन | तिसे | |
| नहीं | घर | वर्ष | दौरान | निकट | यत | |
| पर | समझ | रात | वहीं | छह | एलन | |
| रह | चाहिए | सबसे | सुबह | आर | किर | |
| भी | रूप | कैसे | वर्ग | ओ | जर | |
| कि | जैसा | माँ | डर | पृथ्वी | डल | |
| तो | पहले | बारे | पढ | बज | पडा | |
| ले | बार | आगे | ए | हल | रत | |
| एक | कभी | भाग | तुम्हारा | छत | अल | |
| दे | अच्छा | पी | मा | थक | गर | |
| ही | बोल | अलग | सच | दोपहर | चकमक | |
| ने | कारण | कहाँ | मतलब | तहत | उम | |
| अपना | ओर | विकास | मानो | पे | चन | |
| जो | हाथ | प्राप्त | माध्यम | ओह | दक | |
| आ | कौन | कार्य | अंत | तस्वीर | नक | |
| कह | आज | जगह | उधर | जादू | सकत | |
| कोई | पास | ऊपर | कब | बिंदु | बर | |
| हम | पूरा | शब्द | कुल | सन | आत | |
| सक | वहाँ | बस | जबकि | शक | आद | |
| आप | अधिक | ज्यादा | संख्या | असल | ईस | |
| कुछ | भर | भाषा | एस | धन्यवाद | तया | |
| देख | सुन | बदल | ना | एल | खक | |
| बात | द्वारा | नीचे | हवा | आह | सबस | |
| साथ | देश | बंद | परिवर्तन | लय | रण | |
| क्या | बता | मर | सर | सहमत | मक | |
| दो | क्यों | काफी | बहन | पहल | करत | |
| तक | सारा | खुद | सोना | नफरत | यन | |
| ऐसा | प्रकार | बड़ा | वजह | मुझको | उद | |
| लग | बडा | पिता | मात्र | नरक | साबुत | |
| चल | नया | शहर | मदद | दुबारा | अर | |
| सब | निकल | दुनिया | प्रकाश | गत | यर | |
| बन | लिख | विशेष | खबर | सेट | लन | |
| लोग | पानी | जितना | आग | जनरल | षण | |
| मिल | इसलिए | उपयोग | पद | वर | एसे | |
| या | एवं | अंदर | अत | अप | क्यूंकि | |
| फिर | कई | खेल | बह | दोनो | पत | |
| वाला | अभी | लगभग | आराम | नह | वत | |
| लिए | अगर | स्वयं | रस | आध | रद | |

## IV. Conclusion

At the onset, we set out to create a corpus consisting of aesthetic texts in Hindi and to study the stop words present in this corpus as well as other existing corpora and publicly available lists, with the intent of finding a comprehensive list of stop words. Through this study, we have achieved the following broad objectives:

*a)* To create an unbiased time-independent aesthetics corpus in Hindi

*b)* To study the relationship between the part of speech of a word and its rank in the frequency list

*c)* To provide an exhaustive stop lemma list in Hindi

In order to achieve the first objective, we used texts belonging to fictional and non-fictional categories.

An aesthetics corpus consisting of approximately 1000 articles and 145508 words was built using text from various sources. The gender distribution of the authors was discussed. We could not find significant amount of work by female authors that is available in digital format for public consumption. The corpus consists of both, contemporary texts as well as stories that are over a hundred years old. In order to create an exhaustive stop word list, texts were collated from seven different corpora. The stop word lists that are available for public use were also used. Stop word lists were generated from each corpus based on the word count. It was found that there was an inconsistency even in the top ten stop words across all the lists. We also studied the individual lists and corpora to determine whether the part of speech of a word could be used to determine whether the word can be considered to be a stop word but the results contradicted our assumption. Owing to this observation, while creating a comprehensive list, we did not focus on the part of speech as a characteristic of a word to deem it as a stop word. Only the frequency of the word was taken into count. Through our study, we propose the concept of 'stop lemma' list, instead of mere 'stop word' list, as it is believed to yield better results when used in NLP tasks. Therefore in order to make the list robust, the lemmas of the words were considered as it was found that stop lemmas were more consistent than stop words across different corpora, as can be observed from the density of the respective word clouds generated using the top ten words and the top ten lemmas in the sources . At the time of writing this paper, such a list does not exist in the public domain. This list is limited to the texts collected from the specified sources. We assessed the quality of the list by translating, lemmatizing and comparing the English stop words present in NLTK (a Python package that is widely used in English NLP tasks), with the lemmas from our list. We chose English owing to the familiarity with the language as it helped in determining the quality of translation. We found that almost every lemma that was present in the English list mapped to a corresponding lemma in the Hindi list. Future work arising from this study could comprise evaluation of the list by using it with different NLP tasks, as well as a comparison of the results of a specific NLP task with respect to the use of stop words and stop lemmas in various languages. The corpus, the metadata as well as the analysis has been made available for public consumption at https://github.com/ gayatrivenugopal/hindi-corpus-stoplemmas.

## Acknowledgment

This study is funded by Symbiosis Centre for Research and Innovation, Symbiosis International (Deemed University). We would also like to thank the organizations that provided the corpora that were used in the study, and the Stanford NLP group that released the stanfordnlp package for Python which was used to fetch accurate lemmas of the words in Hindi. The "Linguistic Resources" obtained from TDIL have been developed & made available by TDIL, MeitY, Government of India. The resources from CFILT, IIT-Bombay have been downloaded from http://www.cfilt.iitb.ac.in/Downloads.html.

## References

[1] Ethnologue, https://www.ethnologue.com (2019). Accessed October 21 2019.

[2] Joshi, H., Pareek, J., Patel, R., & Chauhan, K. (2012, December). To stop or not to stop—Experiments on stopword elimination for information retrieval of Gujarati text documents. In 2012 Nirma University International Conference on Engineering (NuiCONE) (pp. 1-4). IEEE.

[3] Jha, V., Manjunath, N., Shenoy, P. D., & Venugopal, K. R. (2016, January). Hsra: Hindi stopword removal algorithm. In 2016 International Conference on Microelectronics, Computing and Communications (MicroCom) (pp. 1-5). IEEE.

[4] Silva, C., & Ribeiro, B. (2003, July). The importance of stop word removal on recall values in text categorization. In Proceedings of the International Joint Conference on Neural Networks, 2003. (Vol. 3, pp. 1661-1666). IEEE.

[5] Na, D., & Xu, C. (2015). Automatically generation and evaluation of stop words list for Chinese patents. Telkomnika, 13(4), 1414.

[6] Larkey, L. S., Connell, M. E., & Abduljaleel, N. (2003). Hindi CLIR in thirty days. ACM Transactions on Asian Language Information Processing (TALIP), 2(2), 130-142.

[7] Martynyuk, S. (2003). Statistical Approach to the Debate on Urdu and Hindi (Student Paper).

[8] Rani, R., & Lobiyal, D. K. (2018). Automatic Construction of Generic Stop Words List for Hindi Text. Procedia computer science, 132, 362-370.

[9] Lo, R. T. W., He, B., & Ounis, I. (2005, January). Automatically building a stopword list for an information retrieval system. In Journal on Digital Information Management: Special Issue on the 5th Dutch-Belgian Information Retrieval Workshop (DIR) (Vol. 5, pp. 17-24).

[10] Zou, F., Wang, F. L., Deng, X., Han, S., & Wang, L. S. (2006, April). Automatic construction of Chinese stop word list. In Proceedings of the 5th WSEAS international conference on Applied computer science (pp. 1010-1015).

[11] Hao, L., & Hao, L. (2008, December). Automatic identification of stop words in chinese text classification. In 2008 International conference on computer science and software engineering (Vol. 1, pp. 718-722). IEEE.

[12] Yao, Z., & Ze-wen, C. (2011, March). Research on the construction and filter method of stop-word list in text preprocessing. In 2011 Fourth International Conference on Intelligent Computation Technology and Automation (Vol. 1, pp. 217-221). IEEE.

[13] Alajmi, A., Saad, E. M., & Darwish, R. R. (2012). Toward an ARABIC stop-words list generation. International Journal of Computer Applications, 46(8), 8-13.

[14] Kaur, J., & Saini, J. R. (2016). POS Word Class Based Categorization of Gurmukhi Language Stemmed Stop Words. In Proceedings of First International Conference on Information and Communication Technology for Intelligent Systems: Volume 2 (pp. 3-10). Springer, Cham.

[15] Choy, M. (2012). Effective Listings of Function Stop words for Twitter. arXiv preprint arXiv:1205.6396.

[16] Kaur, J., & Saini, J. R. (2016, March). Punjabi Stop Words: A Gurmukhi, Shahmukhi and Roman Scripted Chronicle. In Proceedings of the ACM Symposium on Women in Research 2016 (pp. 32-37). ACM.

[17] Saini, J. R., & Rakholia, R. M. (2016). On continent and script-wise divisions-based statistical measures for stop-words lists of international languages. Procedia Computer Science, 89, 313-319.

[18] Rakholia, R. M., & Saini, J. R. (2016). Lexical classes based stop words categorization for Gujarati language. In 2016 2nd International Conference on Advances in Computing, Communication, & Automation (ICACCA)(Fall) (pp. 1-5). IEEE.

[19] Raulji, J. K., & Saini, J. R. (2017, January). Generating Stopword List for Sanskrit Language. In 2017 IEEE 7th International Advance Computing Conference (IACC) (pp. 799-802). IEEE.

## Language Resources

LR#1: Wictionary Top 1900. (June, 2018). In *Wiktionary*. Retrieved July 2, 2019, from https://en.wiktionary.org/wiki/Wiktionary:Frequency_lists/Hindi_1900.

LR#2: 1000 Most Common Hindi Words. (n.d.). In *1000 Most Common Words*. Retrieved July 2, 2019, from https://1000mostcommonwords.com /1000-most-common-hindi-words.

LR#3: First 100 High Frequency Words in Hindi. (May, 2012). In *Hindi Language Blog*. Retrieved July 2, 2019, from https://blogs.transparent.com/hindi/first-100-high-frequency-words-in-hindi.

LR#4: Top 1000 Words. (n.d.). Retrieved July 3, 2019, from http://home.iitk.ac.in/~prasant/HindiCorpus/word.html.

LR#5: Piotr, Most Common Words by Language (2019). In *GitHub*. Retrieved July 3, 2019, from https://github.com/oprogramador/most-common-words-by-language.

LR#6: Savand, A, Stop Words (2017). In *GitHub*. Retrieved July 5, 2019, from https://github.com/Alir3z4/stop-words.

LR#7: Stopwords ISO, Stop Words-hi (2016). In *GitHub*. Retrieved July 10, 2019, from https://github.com/stopwords-iso/stopwords-hi.

LR#8: Champion, Jason, Extra Stop Words (2016). In *GitHub*. Retrieved July 5, 2019, from https://github.com/Xangis/extra-stopwords.

LR#9: Jha, Vandana, N, Manjunath, Shenoy, P Deepa, & K R, Venugopal (2018), “Hindi Language Stop Words List”, Mendeley Data, v1 http://dx.doi.org/10.17632/bsr3frvvjc.1.

LR#10: Hindi Stopwords. (n.d.). Retrieved July 10, 2019, from https://www.ranks.nl/stopwords/hindi.

LR#11: Wikimedia Downloads (n.d.). Retrieved July 11, 2019, from https://dumps.wikimedia.org/backup-index.html.

LR#13: The Open Parallel Corpus (n.d.). Retrieved July 12, 2019, from http://opus.nlpl.eu/.

LR#14: CFILT Hindi Corpus (n.d.). Retrieved July 15, 2019, from https://www.cfilt.iitb.ac.in/Downloads.html.

LR#15: Kunchukuttan, A., Mehta, P. & Bhattacharyya, P. (2018). The IIT Bombay English-Hindi Parallel Corpus. Language Resources and Evaluation Conference.

LR#16: English-Hindi Tourism Text Corpus – EILMT (October, 2016). EILMT Consortia, CDAC Pune. Retrieved July 15, 2019, from http://www.tdil-dc.in.

LR#17: Hindi-English Agriculture & Entertainment Text Corpus ILCI-II (May, 2017). ILCI Consortium, JNU. Retrieved July 15, 2019, from http://www.tdil-dc.in.

LR#18: Hindi Monolingual Text Corpus ILCI-II (June, 2017). ILCI-II, JNU. Retrieved July 15, 2019, from http://www.tdil-dc.in.

LR#19: Hindi-English Health Text Corpus-ILCI (April, 2012). ILCI Consortium, JNU. Retrieved July 15, 2019, from http://www.tdil-dc.in.